# Fully Parallel Architecture for Semi-global Stereo Matching with Refined Rank Method

Yiwu Yao and Yuhua Cheng, *Fellow*, *IEEE*
Shanghai Research Institute of Microelectronics (SHRIME), Peking University, Shanghai 201203, China
School of Electronics Engineering and Computer Science, Peking University, Beijing, 100871, China
Email: yiwuyao@pku.edu.cn

*Abstract*—Fully parallel architecture at disparity-level for efficient semi-global matching (SGM) with refined rank method is presented. The improved SGM algorithm is implemented with the non-parametric unified rank model which is the combination of Rank filter/AD and Rank SAD. Rank SAD is a novel definition by introducing the constraints of local image structure into the rank method. As a result, the unified rank model with Rank SAD can make up for the defects of Rank filter/AD. Experimental results show both excellent subjective quality and objective performance of the refined SGM algorithm. The fully parallel construction for hardware implementation of SGM is architected with reasonable strategies at disparity-level. The parallelism of the data-stream allows proper throughput for specific applications with acceptable maximum frequency. The results of RTL emulation and synthesis ensure that the proposed parallel architecture is suitable for VLSI implementation.

*Keywords—stereo matching; SGM; Rank SAD; fully parallel architecture*

## I. Introduction

As one of the most emphatically studied topics in computer vision, binocular stereo matching is mainly investigated to reconstruct the three-dimensional (3-*D*) geometry (as shown in Fig. 1) from the image pairs with the two-dimensional (2-*D*) spatial or temporal correlation information (intensity, color or texture etc.), which has been extensively applied in the fields including robot vision, driver assistance system, 3-*D* medical image reconstruction etc.

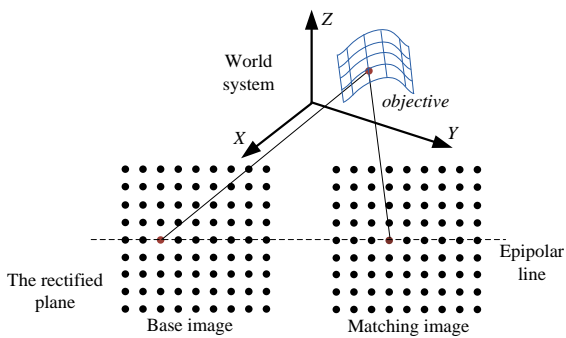

Fig. 1. 3-D geometry reconstruction based on binocular stereo matching: the two images are rectified with camera parameters

The algorithm implementation of binocular stereo matching can be practically classified into local matching method, global optimization, dynamic programming (*DP*) and cooperative matching method [1]. Among the top performing algorithms is the SGM, which is a cooperative algorithm between the local matching and global optimization and possesses the advantages of *DP* method. In order to reduce the complexity of 2-*D* global optimization, the global energy minimization of SGM is firstly decomposed into multiple 1-*D* path cost calculations. Then the best disparity corresponding to the minimum aggregation cost will be determined in a "winner-take-all" manner. Meanwhile, in the decision process along the scan-line of SGM, the inter-consistency of neighboring scan-lines is enhanced, so that the "streaking effect" is effectively suppressed.

Due to the advantageous performance of SGM method, this paper focuses on the research of the refinement of SGM and relevant hardware implementation strategies. The algorithm is refined at the correspondence matching phase by improving the non-parametric rank method (Rank filter/AD). The Rank SAD (RSAD) is proposed as a novel definition for non-parametric correspondence matching, and is used jointly with the Rank filter/AD as a unified rank model to make up for the defects of Rank filter/AD. Consequently, the combination of these two rank methods can help to derive the disparity map with low error matching ratio. The hardware construction for the refined SGM is architected and designed with fully parallel strategies, which is sufficient for real-time implementation targeted for stereo vision applications.

The remainder of this paper is organized as follows. Section II describes the SGM algorithm with refined rank method in detail. Section III presents the hardware architecture of SGM and the related results of RTL emulation and synthesis. Section IV gives the conclusion.

## II. Semi-global Stereo Matching (SGM) with Refined Rank Method

### A. Overview of the SGM Algorithm

As illustrated in Fig. 2, the framework of the SGM stereo matching can be partitioned into four steps: 1) correspondence matching; 2) path cost calculation and cost aggregation; 3) disparity estimation and validation check; 4) refinement of the disparity map. The input image pairs (left / right images) are rectified at the pre-processing stage, and the estimated disparity map can be refined by post-processing with image interpolation, mask filtering or peak removal etc.

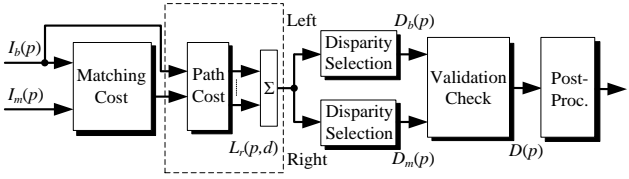

Fig. 2. Framework of the SGM with rank method

Correspondence matching is a major step in most stereo vision applications. Pixel-based matching costs [2] commonly include parametric matching costs (SAD, BT, BilSub/BT, NCC, etc.), non-parametric matching costs (Rank filter/AD, Census filter/Hamming distance, etc.) and the mutual information cost etc. In this paper, the rank-based non-parametric cost is focused to ensure the robustness of the SGM with high performance.

As distinct from other stereo methods, SGM with resort to the global consistency constraints by aggregating path costs along several independent, 1-$D$ paths across the image [3], can effectively improve the estimation accuracy and suppress the "streaking effect" introduced by $DP$. Fig. 3 shows the possible combinations of 1-$D$ paths across the image, where the one illustrated in (c) with the performance similar to (b) on the one hand has the lowest complexity, on the other hand can avoid the causality problems. Therefore, for the target of real-time hardware implementation, the path combination of (c) is the ideal choice for SGM stereo matching.

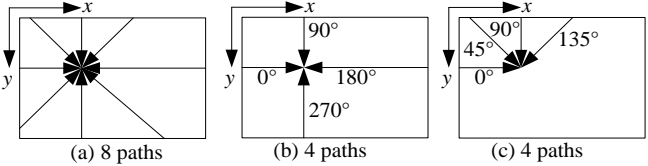

Fig. 3. Possible combinations of 1-$D$ paths for SGM

The path cost $L_r(p, d)$ of pixel $p$ $(x, y)$ along the direction $r$ (0°, 45°, 90°, or 135°) is defined recursively as the following equation:

$$\begin{aligned} L_r(p,d) = C(p,d) + \min\{ & L_r(p-r,d), \\ & L_r(p-r,d-1) + P_1, \\ & L_r(p-r,d+1) + P_1, \\ & \min_i L_r(p-r,i) + P_2\} - \min_k L_r(p-r,k) \end{aligned} \quad (1)$$

where, the data term $C(p,d)$ describes the correspondence cost, and the remaining terms function as the smoothing constraints with the penalty $P_1$ for disparity changes and $P_2$ for disparity discontinuities respectively. Discrimination of small changes ($|\Delta d|=1$) and discontinuities ($|\Delta d|>1$) not only can smooth the slanted or curved surfaces, but also can preserve disparity discontinuities at edges. $P_1$ is an empirically constant, while $P_2$ is adapted to the image gradient $P_2 = P_2' / |I_p - I_{p-r}|$ ($P_2 \geq P_1$).

### B. Refinement of the Non-parametric Rank Method

The SGM stereo methods with SAD, BT, Rank filter/AD are implemented on Visual Studio platform respectively, by utilizing C/OpenCV language. Fig. 4 shows the left/right image pair (Cones dataset) employed for evaluating the performance and efficiency of the SGM algorithm with different matching costs. The disparity is estimated within the range in [0, $D$-1] ($D$=64), which will be scaled to the range in [0, 255] by using quadratic interpolation for rendering after validation check, and will be further median filtered (5x5) for noise reduction at the post-processing stage

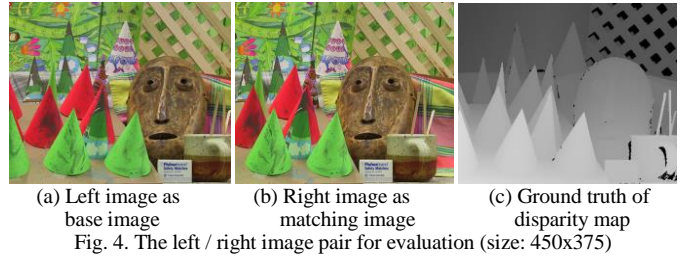

(a) Left image as base image  (b) Right image as matching image  (c) Ground truth of disparity map

Fig. 4. The left / right image pair for evaluation (size: 450x375)

Fig. 5 shows the resulting disparity maps of the SGM with SAD, BT, Rank filter/AD respectively, where the experiment is conducted with no consideration of radiometric differences. In respect to subjective effects, BT method [4] that computes the sampling-insensitive AD between the extreme values of linear interpolations of the corresponding pixels, can achieve the best estimation result. In contrast, the SAD accumulating the ADs within the kernel as matching cost possesses stronger information constraints of neighboring pixels. However, in the absence of sub-pixel accuracy, SAD obtains worse subjective effects than BT. The rank filter can increase the robustness of window-based correspondence matching methods to outliers which usually occur near depth discontinuities and result in blurred borders, but it is sensitive to noise in the texture-less regions attributed to the lack of image structure information as constraints. The time complexity of the SGM with BT that considers less information constraints of neighboring pixels is O($WHMD$), and the time complexity of the other two SGM methods is O($WHMND$) (image size: $W$x$H$, kernel size: $M$x$N$).

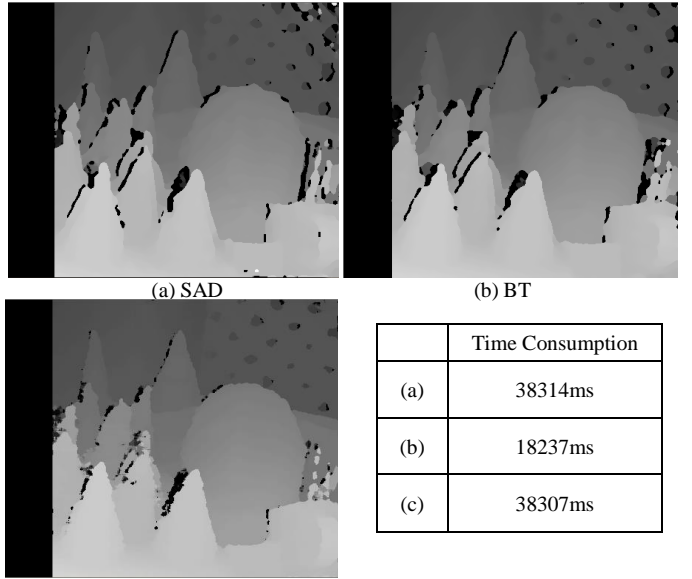

(a) SAD  (b) BT

(c) Rank filter/AD

|   | Time Consumption |
|---|---|
| (a) | 38314ms |
| (b) | 18237ms |
| (c) | 38307ms |

Fig. 5. The resulting disparity maps: the kernel of the correspondence matching is 9x9; and the CPU platform is Intel Core i5-3210M.

Rank filter/AD as a non-parametric matching method has a higher robustness than the parametric method in most cases of radiometric differences (gamma change, etc.) and noise [5]. In order to overcome the deficiency of Rank filter/AD, the joint utilization of SAD and Rank filter/AD is formulated to enhance the constraints of image structure as follow:

$$C(p,d) = \alpha C_{SAD} + (1-\alpha) C_{RT} \quad (2)$$

where $C_{SAD}$ is the matching cost from SAD, $C_{RT}$ is the matching cost from Rank filter/AD, and $\alpha$ is the weighting coefficient. However, that the introduction of parametric approach (SAD) into the non-parametric method will undoubtedly reduce the robustness of stereo matching, and the selection of weighting coefficient is susceptible to radiometric differences and noise.

Referring to the concept of SAD, the Rank filter/AD is re-defined as Rank SAD (RSAD), which is a new non-parametric method with the information constraints of image structure as:

$$C_{RSAD} = \sum_{q \in N_p} \left\{ \left| T\left[I_L(q) < I_L(p)\right] - T\left[I_R(q-d) < I_R(p-d)\right] \right| \right\} \quad (3)$$

where the operator $T[]$ returns 1 if its argument is true and 0, otherwise. According to the equation (3), the Rank SAD is the methodological combination of SAD and Rank filter/AD, so that it possesses strong robustness while considering certain information of local image structure. Furthermore, both rank methods are accomplished relying on rank operation, therefore they can be numerically combined as a unified rank model:

$$C_R = C_{RT} + C_{RSAD}$$
$$= \left| \sum_{q \in N_p} T\left[I_L(q) < I_L(p)\right] - \sum_{q \in N_p} T\left[I_R(q-d) < I_R(p-d)\right] \right| \quad (4)$$
$$+ \sum_{q \in N_p} \left\{ \left| T\left[I_L(q) < I_L(p)\right] - T\left[I_R(q-d) < I_R(p-d)\right] \right| \right\}$$

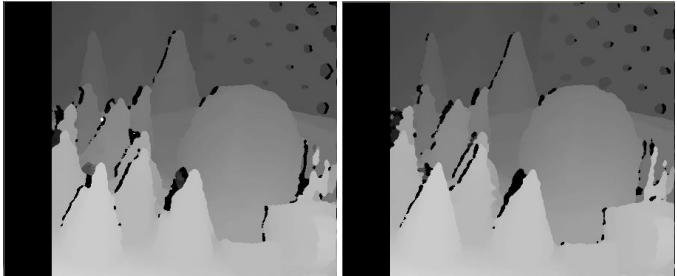

(a) Combination of $C_{SAD}$ and $C_{RT}$    (b) Rank SAD

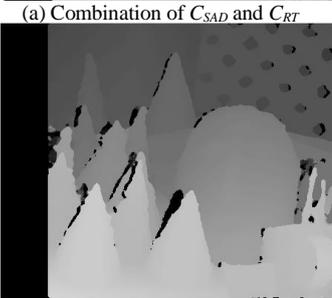

|     | Time Consumption |
| --- | --- |
| (a) | 38502ms |
| (b) | 38470ms |
| (c) | 38781ms |

(c) Combination of $C_{RSAD}$ and $C_{RT}$

Fig. 6. The resulting disparity maps: (a) SAD is introduced to enhance the constraints of image structure ($\alpha$=0.2); (b) Rank SAD is a refined Rank method; (c) $C_{RSAD}$ and $C_{RT}$ are combined as a unified model for stereo matching.

As shown in Fig. 6, the disparity map of (a) is sharper after introducing the role of SAD with certain information of image structure. Meanwhile, due to the enhanced structure constraints and weakened smoothing constraints, the estimation result of (b) reveals better subjective effects than (a), especially in the texture regions or at the object borders. The smoothing effect of Rank filter/AD is retained in the unified rank model, which exhibits certain benefits in (c). The time complexity of these three SGM methods is O($WHMND$), resulting in similar time consuming on CPU platform.

For objective evaluation, the error matching ratio is defined as the following equation:

$$P = \frac{1}{N} \sum_{(x,y)} \left( \left| D_{Calc}(x,y) - D_{Truth}(x,y) \right| > \delta \right) \quad (5)$$

where $\delta$ is the error tolerance, and $\delta$ is set to 1 in this paper. The error matching ratios of the SGM algorithms with different matching costs are summarized in Table 1. It is obvious that the significance of Rank SAD is dramatic for improving the performance of SGM. Meanwhile, the unified rank model with proper smoothing constraints offers the performance similar to the Rank SAD method, so it can be appropriately employed as correspondence matching for hardware implementation of the SGM scheme shown in Fig. 2.

TABLE I. ERROR MATCHING RATIOS OF THE SGM METHODS WITH DIFFERENT MATCHING COSTS

|  | *Cones* | *Teddy* | *Sawtooth* | *Venus* |
| --- | --- | --- | --- | --- |
| SAD | 9.1% | 13.7% | 4.8% | 3.3% |
| BT | 7.7% | 10.4% | 4.0% | 3.1% |
| Rank filter/AD | 7.5% | 11.7% | 3.1% | 5.0% |
| Rank SAD | 6.2% | 9.9% | 2.8% | 2.5% |
| Combination of $C_{SAD}$ and $C_{RT}$ | 8.0% | 12.6% | 3.9% | 3.4% |
| Combination of $C_{RSAD}$ and $C_{RT}$ | 6.6% | 11.1% | 3.2% | 3.8% |

### III. HARDWARE IMPLEMENTATION OF THE SGM ALGORITHM WITH UNIFIED RANK MODEL

#### A. Fully Parallel Architecture of Proposed Algorithm

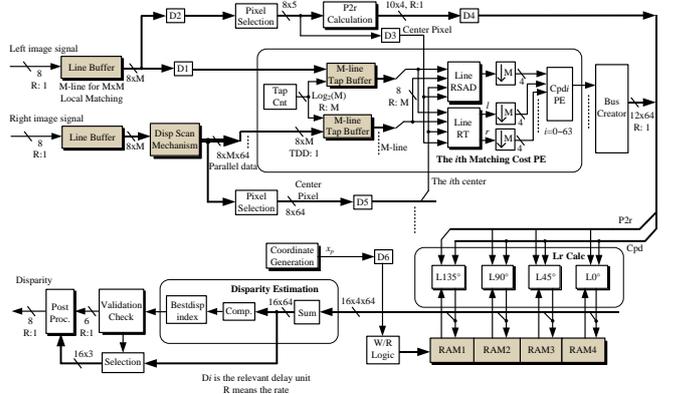

Fig. 7. Fully architecture at disparity-level for hardware implementation of the SGM algorithm with unified rank model

Fig. 7 illustrates the fully parallel architecture at disparity-level for hardware implementation of the SGM with unified rank model. The proposed construction incorporates two 9-line buffers for correspondence matching within 9x9 area, the P2r calculation module relying on LUT, the matching cost PE based on unified rank model, the $L_r$ calculation module with on-chip buffer, the disparity estimation module, the validation check module, and the post-processing module, etc.

The fully architecture strategies are exploited to ensure the performance and efficiency in real-time applications. The 9x9 window-based correlation matching is split into 9 parallel line-based matching operations. The line-based Rank filter/AD and Rank SAD execute in parallel at each line-based matching cost PE. Furthermore, at the disparity-level, the data-stream of the overall structure is divided into 64 parallel processing paths for matching-window buffering, correspondence matching and $L_r$ computation. Hence, the maximum frequency of the parallel scheme is $f_{max} = 9f_S$ ($T_S = 1/f_S$ is the sampling period of the input image signal). Table 2 shows the operating frequencies for applications with specific requirements of throughput.

TABLE II.  MAXIMUM FREQUENCIES FOR DIFFERENT APPLICATIONS

|  | *450x375/ 30fps* | *450x375/ 60fps* | *640x480/ 30fps* | *640x480/ 60fps* | *720p/ 30fps* |
|---|---|---|---|---|---|
| $f_{max}$/MHz | 45.6 | 91.1 | 82.9 | 165.9 | 248.8 |

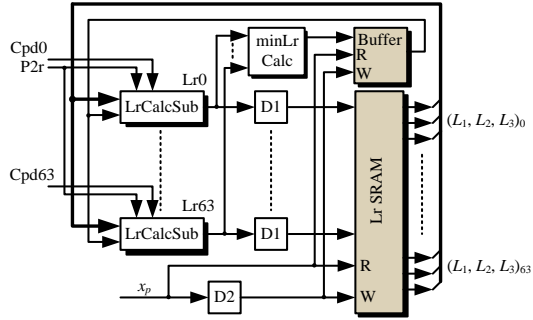

Fig. 8. Diagram of the $L_r$ calculation module with on-chip buffering mechanism for efficient parallel computation at disparity-level

The path costs of pixel $p(x, y)$ along the direction of 0°, 45°, 90°, and 135° are calculated in parallel by reusing the identical computation engine. Fig. 8 illustrates the reusable structure for $L_r$ computation, which consists of 64 sub-PE units executing in accordance with equation (1), the minLr unit for searching the minimum path cost by utilizing comparator tree, the on-chip memory buffer, and several delay units for timing coordination. Note that the signal P2r, $C_{pd}$, minLr, $L_1$, $L_2$ and $L_3$ should simultaneously reach to the relevant sub-PE unit. The $L_1$, $L_2$ and $L_3$ represent the following data terms:

$$\begin{cases} L_1 = L_r(p-r,d) \\ L_2 = L_r(p-r,d-1) + P_1 \\ L_3 = L_r(p-r,d+1) + P_1 \end{cases} \quad (6)$$

Additionally, the usage of on-chip memory for buffering the $L_r$ calculation results is:

$$M_b = \left[(D+5)(3W+1)\right]\left\lceil \log_2(L_{max}+1) \right\rceil \text{ bits} \quad (7)$$

## B. Results of the RTL Emulation and Synthesis

Fig. 9(a) shows the RTL emulation result of the proposed parallel architecture, which is better than the result presented in [6]. Table 3 concludes the synthesis result on Xilinx FPGA platform and the synthesis result by using Design Compiler with GSMC 0.18ULL process for 450x375/60fps application.

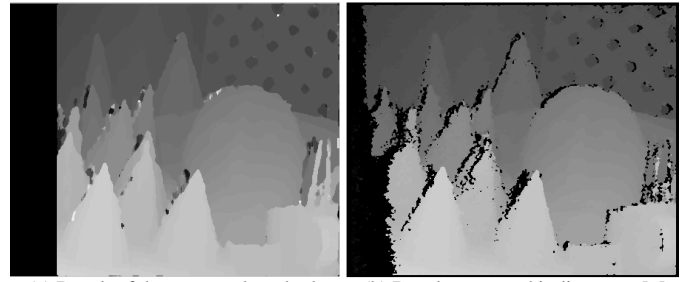

(a) Result of the proposed method  (b) Result presented in literature [6]
Fig. 9. Emulation result of the proposed parallel architecture with 5x5 median filter and the result in the reference literature [6]

TABLE III.  RESULTS OF RTL SYNTHESIS OF THE PROPOSED STRUCTURE

| Process Technology | GSMC 0.18ULL | Supply Voltage | Core 1.8V, I/O 3.3V |
|---|---|---|---|
| Logic Cell Size | 14.93 mm² | On-chip SRAM | 183KB |
| Xilinx FPGA | Synthesized Logic Resource | | |
|  | Slice Registers | Slice LUTs | |
| Spartan 6 xc6slx150 | 13084 / 54576 | 31495 / 92152 | |

## IV. CONCLUSIONS

This paper describes the fully parallel architecture strategies for hardware implementation of the refined SGM with unified rank model. The proposed unified rank model with Rank SAD can make up the deficiency of Rank filter/AD, which leads to favorable results on subjective and objective evaluation. The RTL realization of the SGM is parallelized at disparity-level, where the line correspondence matching is operated by fast time division multiplexing. As a result, the parallel architecture for the SGM with unified rank model is suitable for real-time stereo vision applications.